%% file: root.tex
\documentclass[letterpaper, 10 pt, conference]{ieeeconf}  

\IEEEoverridecommandlockouts                              

\input{preamble.tex}
\usepackage[utf8]{inputenc}
\usepackage{commath}
\usepackage{mathtools}
\usepackage{mathabx}
\usepackage{siunitx}
\usepackage[ruled,vlined]{algorithm2e}
\usepackage[lofdepth,lotdepth]{subfig}

\usepackage[belowskip=-10pt,aboveskip=0pt]{caption}
\graphicspath{{media/}}

\newcommand{\modifMax}[1]{\textcolor{Black}{#1}}

\newcommand{\SO}{\mathrm{\mathbf{SO}}}

\newcommand{\so}{\mathfrak{so}}

\newcommand{\C}{\bm{C}}
\newcommand{\bepsilon}{\bm{\epsilon}}
\newcommand{\bphi}{\bm{\phi}}

\newcommand{\bbeta}{\bm{\beta}}

\newcommand{\bSigma}{\bm{\Sigma}}
\newcommand{\be}{\bm{e}}

\newcommand{\J}{\bm{J}}

\acrodef{lidar}{Light Detection And Ranging}
\acrodefplural{lidar}{Light Detection And Ranging}
\acrodef{ICP}{Iterative Closest Point}
\acrodef{GNSS}{Global Navigation Satellite System}
\acrodef{IMU}{Inertial Measurement Unit}
\acrodef{DOF}{Degrees Of Freedom}

\title{\LARGE \bf
Improving the Iterative Closest Point Algorithm using Lie Algebra
}

\author{Maxime Vaidis$^{1}$, Johann Laconte$^{{1,2}}$, Vladim\'ir Kubelka$^1$ and François Pomerleau$^1$
\thanks{*This work was not supported by any organization}
\thanks{$^{1}$Northern Robotics Laboratory, Université Laval, Québec City, Québec, Canada, $\{$maxime.vaidis, francois.pomerleau$\}$@norlab.ulaval.ca }%
\thanks{$^{2}$Universit\'e Clermont Auvergne, CNRS, SIGMA Clermont, Institut Pascal, F-63000 CLERMONT-FERRAND, FRANCE; johann.laconte@uca.fr}%
}

\begin{document}

\maketitle
\thispagestyle{empty}
\pagestyle{empty}

\begin{abstract}
Mapping algorithms that rely on registering point clouds inevitably suffer from local drift, both in localization and in the built map.
Applications that require accurate maps, such as environmental monitoring, benefit from additional sensor modalities that reduce such drift.
In our work, we target the family of mappers based on the \ac{ICP} algorithm which use additional orientation sources such as the \ac{IMU}.
We introduce a new angular penalty term derived from Lie algebra.
Our formulation avoids the need for tuning arbitrary parameters.
Orientation covariance is used instead, and the resulting error term fits into the \ac{ICP} cost function minimization problem.
Experiments performed on our own real-world data and on the KITTI dataset show consistent behavior while suppressing the effect of outlying \ac{IMU} measurements.
We further discuss promising experiments, which should lead to optimal combination of all error terms in the \ac{ICP} cost function minimization problem, allowing us to smoothly combine the geometric and inertial information provided by robot sensors.
\end{abstract}

\section{Introduction}
\input{tex/intro}

%
\input{tex/RW}

\begin{figure}[thbp]
    \centering
    \includegraphics[width=\linewidth]{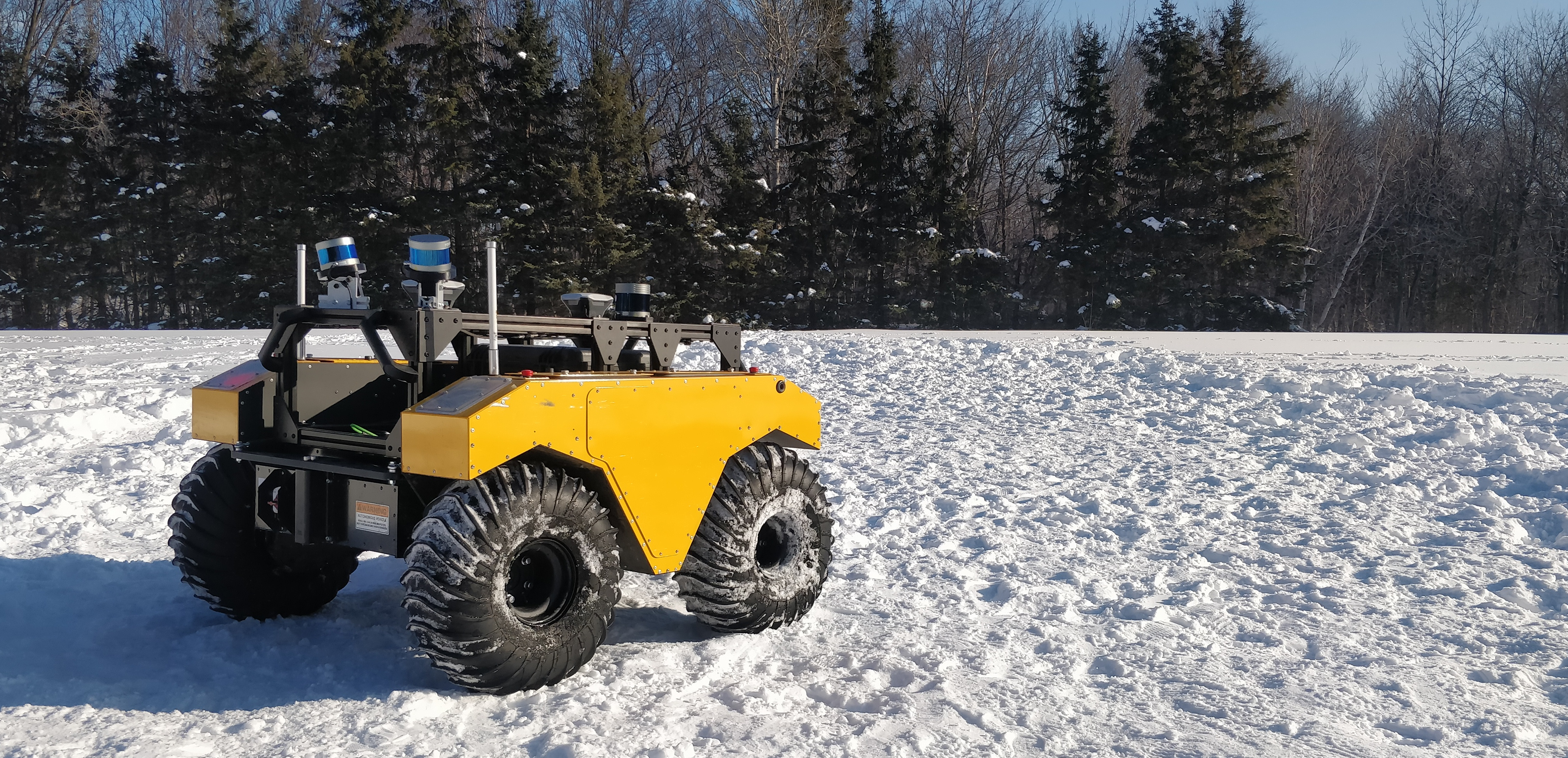}
	\caption{Warthog robot from Clearpath used in the experiments on a snow-covered field. Because of the aggressive maneuvers and the low friction of the snow, the robot produces wrong odometry, leading the \ac{ICP} to fail to converge.}
    \label{fig:intro}
    \vskip-1em
\end{figure}

\section{Theory}
\input{tex/theory}

\section{Experiments}
\input{tex/experimentations}

\section{Conclusion}
\input{tex/conclusion}


\renewcommand*{\bibfont}{\small} 
\printbibliography

\end{document}

%% file: preamble.tex
\usepackage[l2tabu,orthodox]{nag}

\usepackage[
backend=biber, 
bibencoding=utf8,
style=ieee, 
sorting=none, 
natbib=true, 
doi=false, 
isbn=false, 
url=false, 
eprint=false, 
maxcitenames=1, 
mincitenames=1,
]{biblatex}

\usepackage[draft,pdftex,colorlinks]{hyperref}

\usepackage[printonlyused]{acronym}

\usepackage{siunitx}
\usepackage[all]{nowidow}

\usepackage[dvipsnames]{xcolor}

\usepackage{lipsum}

\usepackage[pdftex]{graphicx}

\usepackage{epstopdf}

\usepackage{import}

\graphicspath{{./latexGoodPractices/}}

\usepackage{booktabs}

\usepackage{tabu}

\usepackage{amssymb,amsfonts,amsmath,amscd}

\usepackage{bm} 

\newcommand{\bbm}{\begin{bmatrix}}
\newcommand{\ebm}{\end{bmatrix}}

%% file: tex/intro.tex
In many applications, robots need accurate maps of the environment to complete their tasks.
For instance, our motivation is the problem of environmental monitoring in subarctic areas, as shown in \autoref{fig:intro}, where an autonomous robot needs to reliably locate itself and the measurements it performs.
The \ac{ICP} algorithm is a steady option for registering captured point clouds by minimizing their alignment error.
However, if applied in incremental manner as lidar odometry, it is prone to drift in the estimated pose.
\citet{Babin2019} introduced penalties in the \ac{ICP} cost function suppressing this drift in translation, and also a new point-to-Gaussian minimization.
We will refer to these as Babin's minimization in this paper.
In this work, we extend the concept of penalties of the aforementioned cost function by using Lie algebra which allows us to better constrain the rotation part.

%% file: tex/RW.tex
Our work builds on the results of \citet{Barfoot2014} who adapted tools of the Lie algebra for robotics.
Following \citet{Forster2017} who used the Lie algebra to model uncertainties in rotations for accurate visual odometry, we use the same type of penalties to constrain the \ac{ICP} process, leading to more accurate maps.

%% file: tex/theory.tex
The \ac{ICP} algorithm estimates a rigid transformation $\widehat{\bm{T}}$ that best aligns a set of 3D points $\mathcal{Q}$ (i.e., a \emph{map} point cloud) with a second set of 3D points $\mathcal{P}$ (i.e., \emph{scan} point cloud), given a prior transformation $\widecheck{\bm{T}}$.
Babin's minimization implements the penalties by adding virtual points into $\mathcal{Q}$ and $\mathcal{P}$, steering the \ac{ICP} estimate towards a desired pose.
The pose is provided by a global satellite positioning and an \ac{IMU} for respectively the position and orientation.
We seek the transformation $\widehat{\bm{T}}$ minimizing the penalized distance between the two point clouds such as \autoref{eq:icp_min_with_penalties} where the angular penalty in red is our contribution derived in the next paragraph.
\begin{equation}
\begin{aligned}
\label{eq:icp_min_with_penalties}
\widehat{\bm{T}} = \arg \min_{\bm{T}}  
\underbrace{\frac{\alpha_p}{M} \sum_{m=1}^M w_m\bm{e}^T_m \bm{W}^{-1}_m \bm{e}_m}_{\text{Point clouds}} + 
\alpha_t\underbrace{\bm{e_t}^T \bm{\bSigma_t}^{-1} \bm{e_t}}_{\modifMax{\text{GNSS penalty}}} 
\\ + \alpha_\theta \hspace{0.1cm} \textcolor{BrickRed}{\underbrace{\bm{e_\theta}^T \bm{\bSigma_\theta}^{-1} \bm{e_\theta}}_{\modifMax{\text{Lie penalty}}}}\text{, with \hspace{0.3cm}}
\modifMax{\bm{e}_{m} = \bm{q}_m - \widecheck{\bm{T}} \bm{p}_m,}
\end{aligned}
\end{equation}
\modifMax{where $\bm{p}_m\in\mathcal{P}$ and $\bm{q}_m\in\mathcal{Q}$. The quantities $\bm{e}_{m}$, $\bm{W}_m$ and  $\bm{e}_{t}$ are the error, the covariance of the Gaussians representing the $M$ points of the scan and the GNSS penalty as done by Babin's minimization}. $w_m$ is a weight limiting the impact of outliers.
The covariance matrices $\bSigma_t$ and $\bSigma_\theta$ depict respectively the covariance of the translation and rotation error coming from sensor readings.
Finally, the scaling factors $\alpha_p, \alpha_t, \alpha_\theta$ are all set to $1$ in our work.
Thereby, our contribution lies in a new definition of the penalty on the rotational part.
Our Lie penalty representation does not require manual tuning of parameters without a clear physical meaning, whereas Babin's penalties require virtual points along given axes at a certain distance.

As the translation error is already covered by Babin's minimization, we focus on the formulation of the rotation error $\bm{e_\theta}, \bSigma_\theta$.
For an introduction to the Lie algebra and the notations used in this article, see \citet{Barfoot2014}.
We denote the rotation given by the sensors by $\C_s\in\SO(3)$ and the rotation of the ICP by $\C_I\in\SO(3)$.
Because of the noisy readings, the rotation $\C_s$ is perturbed by a small amount $\bepsilon\in\so(3)$ such that $\bepsilon \sim \mathcal{N}(\bm{0},\bm{\Sigma}_{\bepsilon})$.
We choose to perturb $\C_s=\exp(\bphi^\wedge)$ in the left, meaning that $\C_s = \exp(\bepsilon^\wedge)\bar{\C}_s$.

Using the above definitions, we compute the error as
\begin{equation}
\begin{aligned}
\label{eq:error_equation}
	\exp(\be_\theta^\wedge) 
					 &= \exp(\bepsilon^\wedge)\exp(\bbeta^\wedge),
\end{aligned}
\end{equation}
with $ \exp(\bbeta^\wedge)=\C_s\C_I^T $ being the difference of the rotation between the \ac{ICP} and the sensors reading.
We can then approximate the above equation, leading to
\begin{equation}
\begin{aligned}
\label{eq:error_equation_approx}
 	\exp(\be_\theta^\wedge) &\approx \exp((\bbeta + \J(\bbeta)^{-1}\bepsilon)^\wedge) \\
					 &\Rightarrow \be_\theta \approx\bbeta + \J(\bbeta)^{-1}\bepsilon,
\end{aligned}
\end{equation}
where $\J(\bbeta)$ is the left Jacobian of $\bbeta$ as defined in \cite{Barfoot2014}.
We can then compute the angular penalty term, defined as the Mahalanobis distance of $\bbeta\in\so(3)$ from the origin as
\begin{equation}
	\begin{aligned}
	\label{eq:error_final}
		\bm{e_\theta}^T \bm{\bSigma_\theta}^{-1} \bm{e_\theta} &= \bbeta^T\J(\bbeta)^T\bSigma_\epsilon^{-1}\J(\bbeta)\bbeta 
			= \bbeta^T\bSigma_\epsilon^{-1}\bbeta,
	\end{aligned}
\end{equation}
\modifMax{since $\J(\bbeta)\bbeta = \bbeta$, by developping the Jacobian.}
Therefore, the only parameter to provide is the covariance $\bSigma_\epsilon$ of the sensors' reading rotation error.
In the case where the source of the rotation measurement feeds a bad prior to the \ac{ICP}, meaning that its covariance matrix is large, the penalty will have little effect on the optimization.

%% file: tex/experimentations.tex
To demonstrate the performance of our method, we tested our algorithm in a snowy environment at the Laval University as shown in \autoref{fig:intro}, and with the SLAM evaluation benchmark from the KITTI dataset \cite{Geiger2012}.
\modifMax{The prior takes the GNSS point for the translation, and the IMU for the orientation.}
To ease the tuning of remaining parameters and the data analysis, we constrained the \ac{ICP} to optimize only the position and the yaw angle.
Because of the \modifMax{experiments conditions} (robot on a mostly flat surface, car on a road), we could accept the prior roll and pitch angles as correct.

\begin{figure}[thbp]
	\centering
	\includegraphics[width=\linewidth, trim={0 90 0 90},clip]{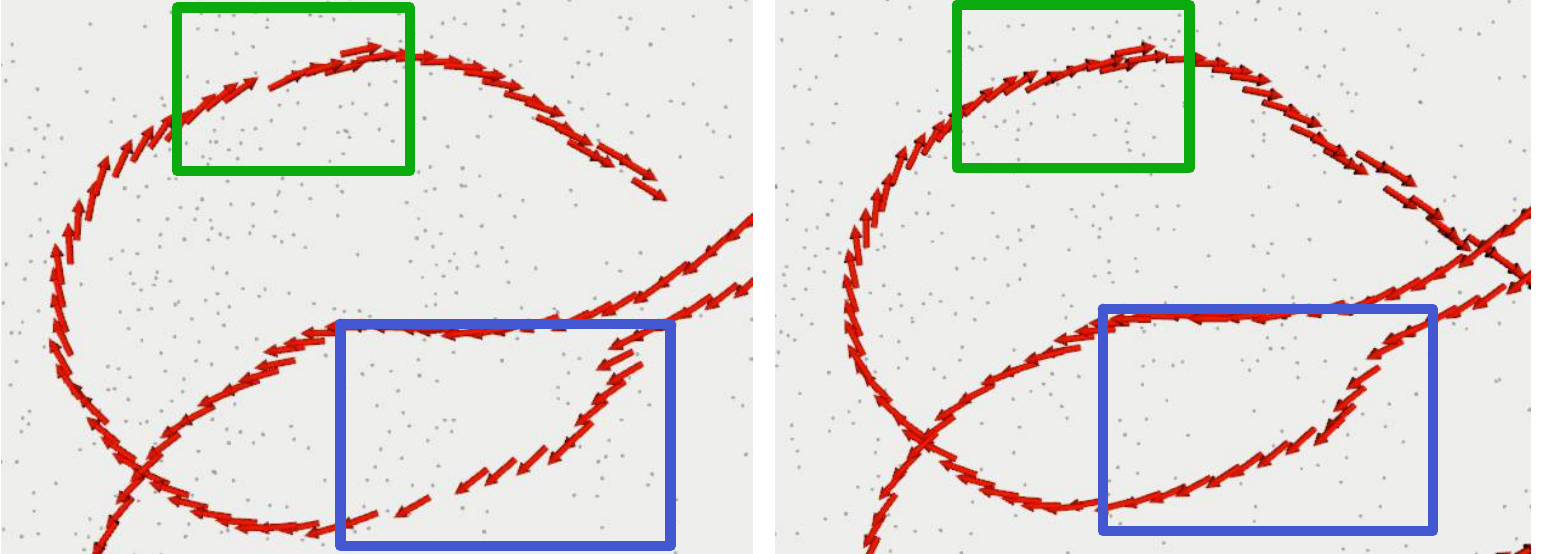}
	\caption{Comparison between the trajectories obtained without (left) and with (right) the rotational penalty. Without the penalty, the \ac{ICP} fails to converge at different positions, leading to "breaks" or holes in the trajectory of the robot (shown in green and blue boxes).}
    \label{fig:Warthog}
\end{figure}

Firstly, the data were gathered using a Warthog from Clearpath Robotics and a RS32 Robosense lidar.
During this experiment, the robot was performing high-velocity trajectories: 
because of the low friction of the snow, the robot was considerably skidding and slipping and thus generating incorrect odometry prior.
Due to the high velocity, the distance between subsequent point clouds can be large and with wrong prior, the \ac{ICP} can fall into local minimum by wandering too far away from the orientation prior.
\autoref{fig:Warthog} shows two computed trajectories for the same experiment, with and without our rotational penalty.
On the trajectory computed without the penalty, two breaking points can be seen at the top and bottom part of the picture, meaning that the \ac{ICP} converged to a wrong solution.
With the Lie penalty, these breaking points are avoided as this term pushes the \ac{ICP} to converge quicker to the right solution while avoiding local minimum.

Then, we further used the SLAM evaluation benchmark from the KITTI dataset \cite{Geiger2012} to compare our results.
In the ten trajectories tested, the dataset 01 shows the best improvements.
With the Lie penalty, we obtained a translation error of \SI{4.88}{\%} and a rotational error of \SI{1.55}{deg/100m} which is an improvement of \SI{0.79}{\%} and \SI{0.41}{deg/100m} compared to the one without the penalty.
\modifMax{In difficult environments with very few geometrical features to match, our Lie penalty guides the minimization by relying on the \ac{IMU}.}
In well-structured ones, the \ac{ICP} can find a lot of features to effectively reduce the residual error between the subsequent point clouds.
Moreover, a car cannot rotate as fast as a skid-steered robot, meaning that the \ac{ICP} has a good prior in most of the cases.
Thereby, the best improvements are observed where the \ac{ICP} does not have a good prior or where it is not constrained enough by the point-cloud geometry.

%% file: tex/conclusion.tex
In this article, we proposed a novel method to penalize the rotation of the \ac{ICP} algorithm with Lie algebra.
Using \modifMax{preliminary} real-world experiments, we showed that our method leads to increased accuracy and stability of the \ac{ICP}.
Future work will focus on the development of these penalties aimed at improving the \ac{ICP} algorithm, while extensively proving their added value to the mapping process.
The Lie penalty will be improved, directly taking into account the noise in the angular rate of the \ac{IMU}.
\modifMax{To provide ground truth reference, the robot will be tracked by three theodolites during others many experiments to obtain accurate comparisons.}
Furthermore, the penalty will take into account the kinematic model of the robot using the work of \citet{Baril2020}.